\documentclass[sigconf]{acmart}
\AtBeginDocument{%
  }

\setcopyright{acmlicensed}
\copyrightyear{2026}
\acmYear{2026}
\acmDOI{XXXXXXX.XXXXXXX}
\acmConference[Conference acronym 'XX]{Make sure to enter the correct
  conference title from your rights confirmation email}{}{}
\acmISBN{}

\usepackage{hyperref} 
\usepackage{float}
\usepackage[capitalize]{cleveref}
\usepackage[linesnumbered,ruled,vlined]{algorithm2e}
\crefname{section}{Sec.}{Secs.}
\Crefname{section}{Section}{Sections}
\Crefname{table}{Table}{Tables}
\crefname{table}{Tab.}{Tabs.}
\crefname{algorithm}{Algo.}{Algos.}
\usepackage{multirow}

\begin{document}

\title{GenMRP: A Generative Multi-Route Planning Framework for Efficient and Personalized Real-Time Industrial Navigation}


\author{Chengzhang Wang}
\authornote{Both authors contributed equally to this research.}
\email{wangchengzhang.wcz@alibaba-inc.com}
\author{Chao Chen}
\authornotemark[1]
\email{cc201598@alibaba-inc.com}
\affiliation{%
  \institution{Amap, Alibaba Group}
  \city{Beijing}
  \country{China}
}

\author{Jun Tao}
\affiliation{%
  \institution{Amap, Alibaba Group}
  \city{Beijing}
  \country{China}}
\email{senbao.tj@alibaba-inc.com}

\author{Tengfei Liu}
\affiliation{%
  \institution{Southern University of Science and Technology}
  \city{Shenzhen}
  \country{China}}
\email{12332470@mail.sustech.edu.cn}

\author{He Bai}
\affiliation{%
  \institution{Amap, Alibaba Group}
  \city{Beijing}
  \country{China}}
\email{baihe.bh@alibaba-inc.com}

\author{Song Wang}
\affiliation{%
  \institution{Amap, Alibaba Group}
  \city{Beijing}
  \country{China}}
\email{wang.song@alibaba-inc.com}

\author{Longfei Xu} 
\authornote{corresponding author}
\affiliation{%
  \institution{Amap, Alibaba Group}
  \city{Beijing}
  \country{China}}
\email{longfei.xl@alibaba-inc.com}

\author{Kaikui Liu}
\affiliation{%
  \institution{Amap, Alibaba Group}
  \city{Beijing}
  \country{China}}
\email{damon@alibaba-inc.com}

\author{Xiangxiang Chu}
\affiliation{%
  \institution{Amap, Alibaba Group}
  \city{Beijing}
  \country{China}}
\email{cxxgtxy@gmail.com}

\renewcommand{\shortauthors}{Chengzhang Wang et al.}

\begin{abstract}
  Existing industrial-scale navigation applications contend with massive road networks, typically employing two main categories of approaches for route planning. The first relies on precomputed road costs for optimal routing and heuristic algorithms for generating alternatives, while the second, generative methods, has recently gained significant attention. However, the former struggles with personalization and route diversity, while the latter fails to meet the efficiency requirements of large-scale real-time scenarios. To address these limitations, we propose GenMRP, a generative framework for multi-route planning. To ensure generation efficiency, GenMRP first introduces a skeleton-to-capillary approach that dynamically constructs a relevant sub-network significantly smaller than the full road network. Within this sub-network, routes are generated iteratively. The first iteration identifies the optimal route, while the subsequent ones generate alternatives that balance quality and diversity using the newly proposed correctional boosting approach. Each iteration incorporates road features, user historical sequences, and previously generated routes into a Link Cost Model to update road costs, followed by route generation using the Dijkstra algorithm. Extensive experiments show that GenMRP achieves state-of-the-art performance with high efficiency in both offline and online environments. To facilitate further research, we have publicly released the training and evaluation dataset. GenMRP has been fully deployed in a real-world navigation app, demonstrating its effectiveness and benefits.
\end{abstract}

\ccsdesc[500]{Applied computing~Transportation}
\ccsdesc[500]{Computing methodologies~Planning and scheduling}
\keywords{Personalized Routing, Alternative Routes Generation, Real-time Link Cost Calculation, Graph Search}

\maketitle

\section{Introduction}
\begin{figure}[t]
  \centering
  \includegraphics[width=0.95\linewidth]{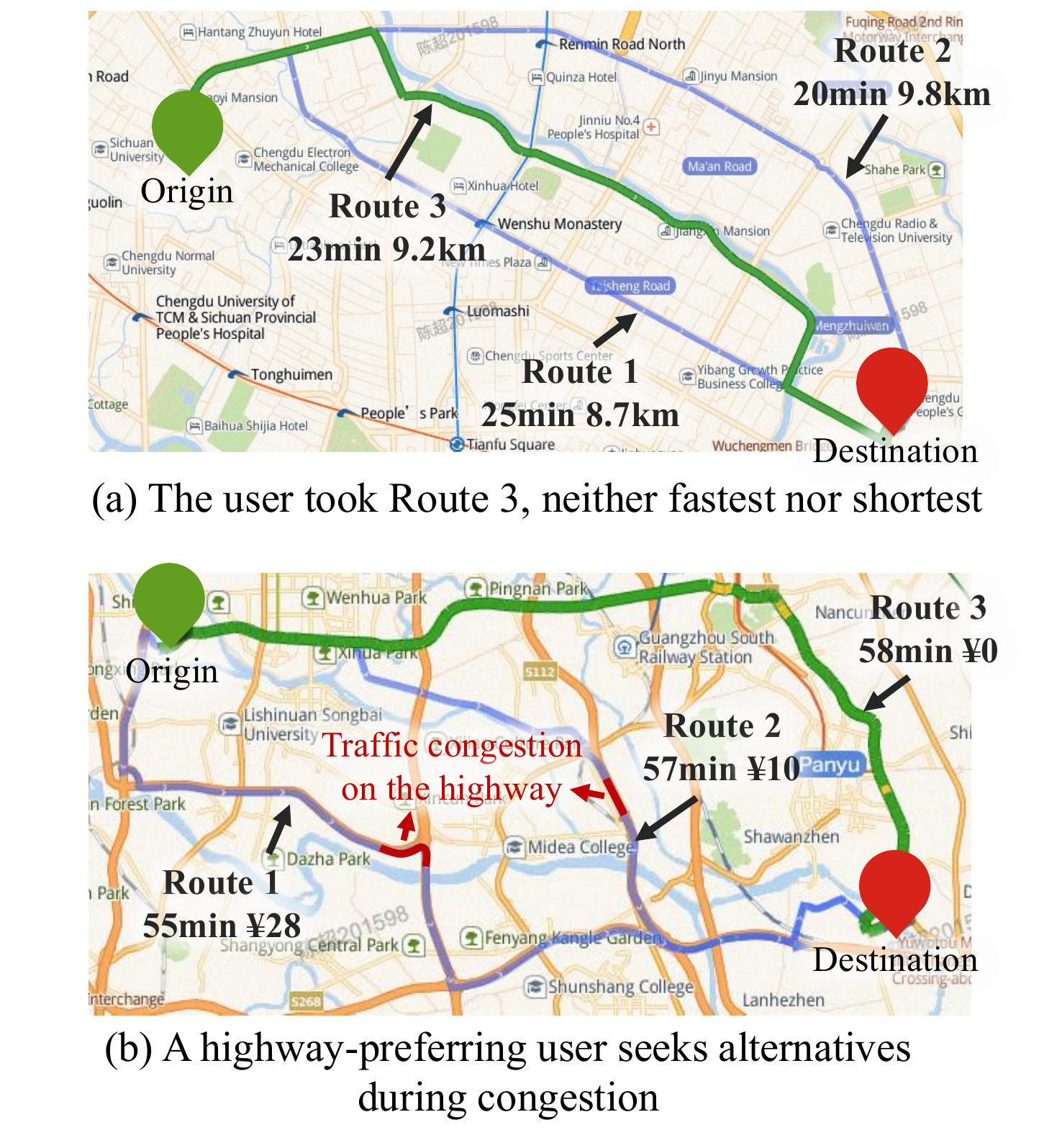}
  \caption{Personalization and diversity are essential for route planning.}
  \label{fig: user_preference}
\end{figure}

Currently, the route planning problem is gaining increasing attention \cite{IJCAI1, WWW1, NIPS1, weighted2009Astar, application2011Astar, most2021diverse, efficient2024alternative, deep2024alternative}, and most studies on route planning focus on optimizing routes based on a limited set of metrics such as travel time or distance. However, in reality, defining a ``good route'' for users is complex. For example, as shown in \cref{fig: user_preference}(a), Route 1 and Route 2 represent the shortest route in distance and time, respectively. However, the user trajectory is Route 3. This indicates that, in addition to standard metrics like time and distance, factors such as route familiarity and real-time congestion significantly shape a user’s evaluation of a route. Furthermore, the same user's preferences can be dynamic across different scenarios. For instance, a user might prioritize time efficiency for their morning commute, yet emphasize drive stability and comfort during long distance family trips. Therefore, personalized and context-awareness route planning is essential.

Beyond personalization, users also have a demand for route comparison. They expect navigation platforms to provide multiple routes with distinct characteristics, offering options to address varying scenarios. As shown in \cref{fig: user_preference}(b), historical data indicate that the user tends to prefer the route with highway (Route 1 or Route 1). However, when highways become congested, non-highway options (Route 3) emerge as critical alternatives for the user's final selection.

Although many studies have explored personalization and alternative route generation \cite{learning2020effective, personalized2022route, efficient2022navigation, ASNN2023FRR, robust2023routing}, three key limitations remain in existing research:

\textbf{Personalization and context-awareness}: Many studies rely on predefined hyperparameters, such as setting thresholds for certain criteria or using explicit weighting functions to calculate the cost of links ('link' mentioned in this paper represents edges in the graph), which are often inflexible when addressing personalized and context-aware needs.

\textbf{Diversity}: Existing methods are mostly based on heuristic algorithms to compute alternative routes but lack effective modeling of diversity to align with user preferences.

\textbf{Computational efficiency}: Achieving personalized, context-sensitive, and diverse multi-route planning requires complex feature representations and deep models, along with real-time processing capabilities. This presents a significant challenge for industrial-grade deployment.

To address these challenges, we propose GenMRP, a \underline{gen}erative \underline{m}ulti-\underline{r}oute \underline{p}lanning framework for efficient and personalized real-time industrial navigation. GenMRP iteratively generates routes with a link cost model (LCM) and a bidirectional Dijkstra routing module (BDR)~\cite{biDij2011}. (1) LCM leverages request-specific contextual information, user historical sequences, user frequency features, and both dynamic and static link features to model personalized and context-aware link costs with DIN \cite{DIN}, GAT \cite{GAT_first}, and Multi-Scenario Network \cite{DSFnet2025}. (2) Diversity is fostered through an iterative process: each iteration feeds the previously generated routes into LCM as conditions and introduces a correctional boosting approach to balance quality and diversity. (3) To boost GenMRP efficiency, we propose the skeleton-to-capillary (STC) method, generating a downscaled sub-road network that preserves critical links for each request. Moreover, during online inference, we update only the links with state changes in each iteration to achieve higher efficiency.

As for route evaluation, we align directly with user trajectories rather than relying on manually defined route metrics such as travel time or distance. User trajectories have been utilized in previous work \cite{RL2019personal, trajectory}, directly aligning the objective with user preferences.

The BDR module makes GenMRP non-differentiable, preventing direct optimization of LCM via gradient descent. To solve this, we designed a sampling-based learning approach: high-quality routes are generated based on multi-dimensional Pareto link costs, with the route most similar to the user's trajectory as the positive sample and the rest as negative samples.

Existing public road network datasets like OpenStreetMap (OSM), Natural Earth (NE) mainly provide basic link features such as travel time, distance, and toll, etc., which are insufficient for modeling complex user preferences. Therefore, we released a comprehensive dataset collected from real navigation applications, containing request-level sub-road networks, dynamic and static link features, user historical sequences, and scenario features. Extensive experiments demonstrate that GenMRP achieves superior performance in both offline and online evaluations. It has been successfully deployed in a real-world online navigation application.

Our main contributions can be summarized as follows.
\begin{itemize}
\item We propose GenMRP, a Generative Multi-Route Planning Framework to enable personalized and context-aware multi-route generation.
\item We propose a correctional boosting approach to achieve route diversity and introduce STC and incremental online inference to improve computational efficiency.
\item Both offline and online experiments validate the effectiveness and efficiency of GenMRP. Moreover, the associated dataset, collected from a real-world navigation application, has been open-sourced, establishing it as the most information-rich public dataset to date in the field of route planning.

\end{itemize}

\section{Related Work}
\subsection{Personalized Recommendation}
Currently, achieving personalized recommendations is a crucial research topic in the field of recommendation systems \cite{rajput2023recommender, zhang2024wukong, ICML3}. In the current field of route personalized recommendation, some researchers have focused on statistically analyzing metrics such as travel time and distance to represent users' preferences. For example, Dai et al. defined user preferences using the ratios of travel time, distance, and fuel consumption \cite{personalized2015route}. Zhu et al. proposed the FineRoute system, a personalized and time-sensitive route recommendation framework \cite{fineroute2017personalized}, which provides fine-grained recommendations using temporal dimensions.

Other researchers rely entirely on historical trajectory data. For example, Sacharidis et al. designated a portion of links in the network as user-preferred routes as prior knowledge \cite{finding2017prefer}. The proportion of preferred routes and the total length were used as metrics to construct a Pareto front, allowing Pareto optimization to balance route quality and personalization. Meanwhile, Silva et al. calculated the transition probabilities of users in the road network based on trajectory data \cite{personalized2022route}, identifying the most likely routes to be chosen by users.

Some studies have explored personalized recommendations for tourist routes. Zhang et al. measured visual and facility diversity using publicly available data such as Google Street View images \cite{walking2018route}. Xu Teng et al. started by identifying key waypoints based on Points of Interest (POI) \cite{semantically2021diverse}, then determined the route, effectively substituting route-level personalization with POI-level personalization.
\subsection{Alternative Routes Generation}
In the field of alternative route generation, most studies still prioritize metrics such as similarity as the primary evaluation criteria. A significant portion of this work builds on the classic k-shortest route problem \cite{most2021diverse, finding2020kshortest, walking2018route, diversified2022topk}. These studies have innovated and optimized algorithms and explored diverse modeling approaches. For instance, Häcker et al. relaxed the strict requirement for ``shortest'' routes and proposed a method to find the k-Most Diverse Near-Shortest Routes.

Penalty addition is another approach \cite{candidate2012set, alternative2013route, alternative2015route}. The core idea is straightforward: routes are generated sequentially, after each route is computed, penalty factors are applied to the links used, ensuring subsequent computations explore alternative routes.

Beyond search-based methods, machine learning approaches have also been explored \cite{deep2020predictive, embedded2020deep, graph2022network, deep2024alternative}. These methods focus on leveraging machine learning to predict auxiliary values that are difficult to obtain directly, such as average speed and traffic flow, rather than explicitly generating routes aligned with user preferences.

\section{Preliminaries}
\subsection{Graph Representation of Road Networks}
\begin{figure}[t]
  \centering
  \includegraphics[width=1\linewidth]{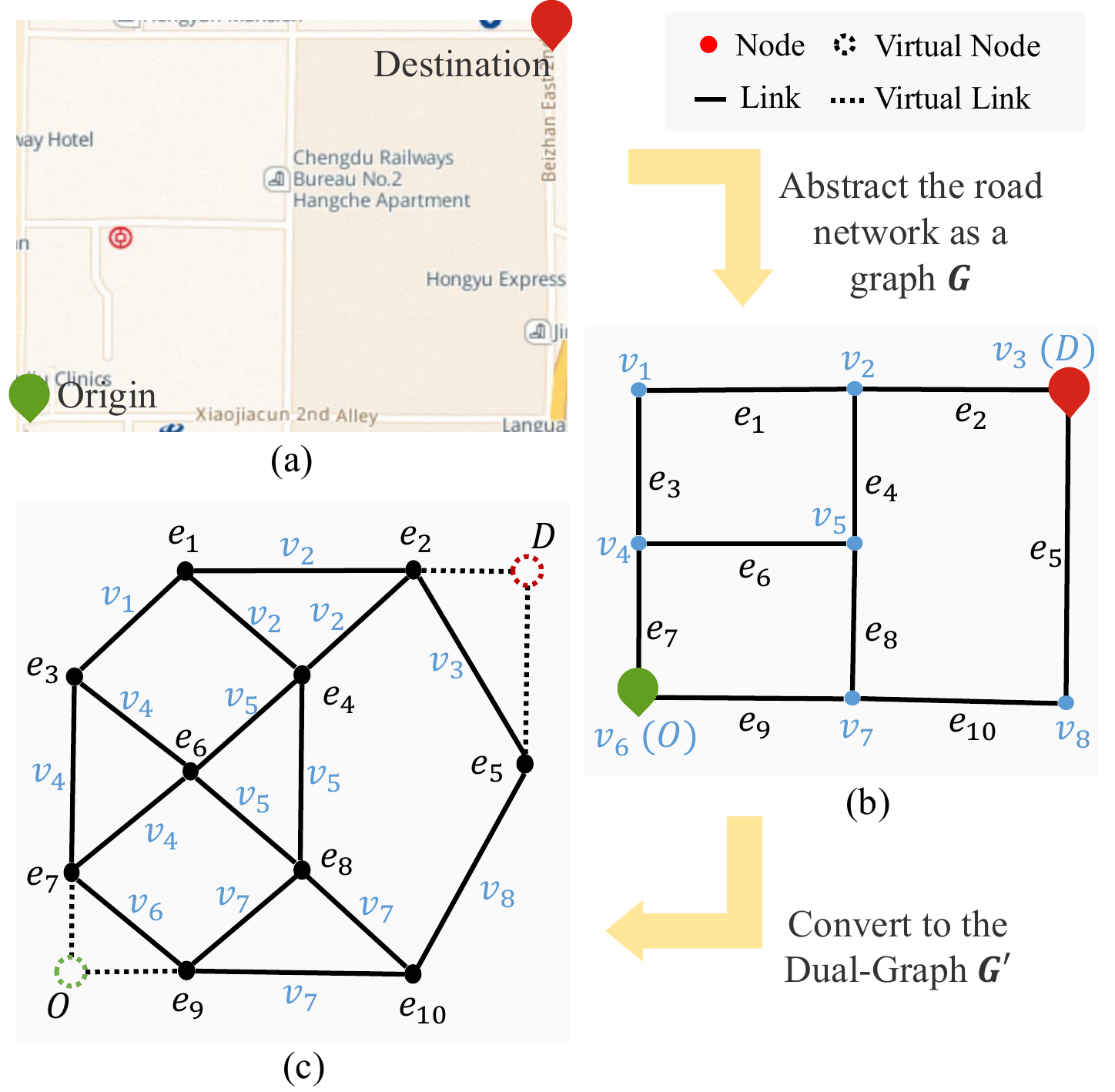}
  \caption{A real-world road network and its corresponding abstracted graph and dual-graph.}
  \label{fig: road network}
\end{figure}

\cref{fig: road network}(a) illustrates a real-world road network comprising multiple intersections and road segments, with the user aiming to travel from the origin (O) to the destination (D).

Intersections are represented as vertices and road segments as edges, abstracting the road network into a graph $G$, as shown in \cref{fig: road network}(b). In real-world scenarios, it is essential to calculate vertex costs along with edge costs since intersections are influenced by attributes like traffic lights, intersection complexity, and turning actions. However, a vertex's turning action depends on its connected upstream and downstream links, varying across routes and making its cost difficult to compute directly. For instance, $v_7$ has different actions: turning left to $e_8$ or going straight to $e_{10}$, which prevents $v_7$ from having a fixed cost.

To address this, we swap the edges and vertices in $G$ to construct the corresponding dual graph $G'$, as shown in \cref{fig: road network}(c). Each edge in $G'$ is defined by an intersection and its two connected road segments, enabling the determination of the intersection's turning action. Further simplify, the edge features in $G'$ are extended to include the features of adjacent vertices, reducing the problem to computing only edge costs.

Based on $G'$, we define link and route as follows:

\textbf{Link}: Represents an edge in $G'$, denoted as $l$. The set of all links in $G'$ is represented as $L = \{l_1, l_2, ..., l_N\}$. $N$ represents the total number of links in the graph.

\textbf{Route}: Represents a path from $O$ to $D$ in $G'$, denoted as $r$. The set of all routes is represented as $R = \{r_1, r_2, ..., r_M\}$. Specifically, the actual trajectory taken by the user is denoted as $r_u$. $M$ denotes the total number of routes in $R$.

\subsection{Metrics}
To align route quality evaluation with user preferences, it is necessary to define a metric related to user trajectories. Following the approach of \cite{RL2019personal}, we use user trajectory coverage to evaluate route quality. The calculation is as follows:

\begin{equation} \label{eq: probForm}
    Cov_i=\frac {|r_u\cap r_i|}{|r_u\cup r_i|} \quad \forall r_i\in R,
\end{equation}

where $|r_u\cap r_i|$ denotes the length of the overlapping links between the two routes, and $|r_u\cup r_i|$ represents the total length of the union of links that compose the two routes. 

We measure whether a route is "good" enough by calculating the coverage of the recommended route because the user's choice of route and driving along the route are the best evaluation indicators representing the user's overall experience of the route. When a large number of users have driven along the route we provided for a long time, we can directly infer that the recommended route has indeed satisfied the users.
    
\subsection{Problem Definition}
After the preceding road network representation and definitions, given $O$, $D$, and $G'$, our objective is to generate $K$ routes to meet the following two properties.

\textbf{Personalization} The goal of personalization is to make the recommended route selectable by users as much as possible. 

\textbf{Diversity} The goal of diversity is to make each recommended route as different as possible from other routes.

In order to achieve these two properties, we propose the correctional boosting objective function for the route generation of the $k$-th iteration:
\begin{equation} \label{eq: obj_f}
    \max \sum_{\Omega}(Cov_{*} - Cov_{k-1}) {P(r_*|O, D, G', R_{k-1})},
\end{equation}
where $\Omega$ is the dataset, $r_*$ denotes the route with the maximum $Cov$ in $R$, $Cov_*$ is the coverage of $r_*$, $R_{k-1}$ is the route set generated before iteration $k$ and $Cov_{k-1}$ is the maximum coverage in $R_{k-1}$. Specially, $R_0=\emptyset, Cov_0=0$.

The growth of the maximum coverage rate of the recommended route set can not only describe the improvement of personalized quality, but also ensure that the level of diversity is not low. In the $k$-th iteration, the learning for personalization is reflected in maximizing the probability of generating $r_*$. Note that only samples where $r_*$ has not been generated have $\Delta Cov_k = Cov_{*} - Cov_{k-1} > 0$. At this stage, the objective function focuses on optimizing these samples, prioritizing those with higher $\Delta Cov_k$. A higher $\Delta Cov_k$ guides the model to explore route generation directions dissimilar to $R_{k-1}$, thereby enhancing diversity. 

According to the definition of coverage, when the similarity between the newly added route and the route in the set is very high, the coverage will not be increased. Therefore, we can know that when the marginal benefit is the highest, the diversity of the route set will not be very low, so as to achieve a balance between the personalized quality of the route and the diversity of the route.

\subsection{The Proposed Dataset}
\begin{table}[t]
  \caption{A Brief Introduction to Dataset PRN’s Features}
  \label{tab: key features}
  \begin{tabular}{lll}
    \toprule
    Name & Dimension & Description \\
    \midrule
    $O$ & $1$ & Origin\\
    $D$ & $1$ & Destination\\
    $G'$ & $N\times 2$ & Link adjacency in $G'$\\
    $x^s$ & $10$ & Contextual features\\
    $x^h$ & $100\times 32$ & User historical sequence\\
    $x^{link}$ & $N\times 50$ & Link features\\
    $x^{freq}$ & $N\times 20 \times 7$ & User Familiarity Frequency\\
    $x^{heat}$ & $N\times2$ & Link heatmap features\\
    $R$ & $M*|r|$ & Link list of each route\\
    $Cov$ & $M \times 1$ & Coverage of each route\\
    \bottomrule
  \end{tabular}
\end{table}

We collected data from a real-world navigation application and constructed the PersonaRoadNet (PRN) dataset. PRN contains approximately 600K samples from about 590k users, each corresponding to a unique route planning request. Each sample includes a specific origin ($O$), destination ($D$), dual graph ($G'$), and a variety of features, as detailed in \cref{tab: key features}. Each sample contains an average of approximately 1300 links. $|r|$ is a variable dimension that represents the length of the link sequence for routes in $R$, with different routes having different values. $x^{link}$, $x^{freq}$, and $x^{heat}$ correspond to link dynamic/static features, user familiarity frequency, and link heatmap features, respectively. Details of these features can be found in Appendix~\ref{Appendix:feature}.

To the best of our knowledge, PRN is the first request-level dataset, and each sample corresponds to a unique user request that is not included in existing public datasets (OSM, NE). In addition, PRN contains a wealth of route network features, including nearly 200 dimensions of dynamic and static features, while OSM and NE can only obtain some simple static features.

\section{Method}
\begin{figure*}[t]
  \centering
  \includegraphics[width=0.9\linewidth]{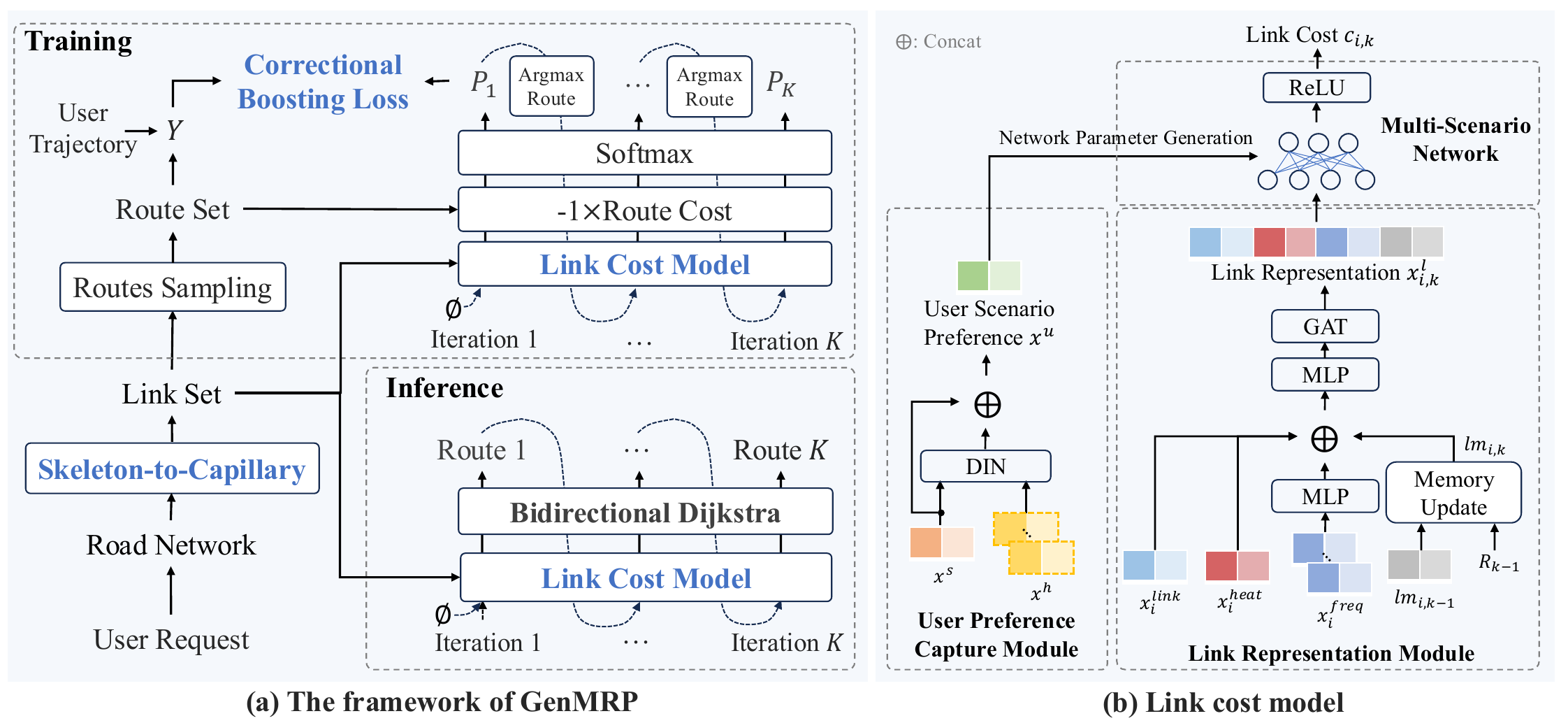}
  \caption{The framework of GenMRP and the Link Cost Model (LCM). The link set are generated based on user requests and the skeleton-to-capillary method. Training involves route sampling, where the link set and previously generated routes are input in each iteration. For inference, the trained link cost model calculates link costs in each iteration, and bidirectional Dijkstra is applied for routing. LCM is designed to compute link costs through its user preference capture module, link representation module, and multi-scenario network. A detailed description of the input features is provided in \cref{tab: key features}.
  }
  \label{fig: framework}
\end{figure*}

GenMRP is a generative multi-route planning framework that processes user requests by iteratively calculating link costs and generating routes. Each iteration utilizes previously generated routes and various features to refine the routing process. The framework is shown in \cref{fig: framework}(a). 

The following sections first introduce the Link Cost Model, followed by the Route Generation and Correctional Boosting method for multi-round route generation. Finally, we detail the techniques for enhancing computational efficiency, including the Skeleton-to-Capillary approach and route sampling strategies.

The Link Cost Model (LCM) is designed to calculate the cost of links in the graph $G'$. As illustrated in \cref{fig: framework}(b), LCM consists of three components:

\textbf{User Preference Capture Module}: This module utilizes the DIN \cite{DIN} model to capture the user's scenario route selection preferences $x^u$ by modeling contextual features $x^s$ and user historical sequences $x^h$.

\textbf{Link Representation Module}: For link $l_i$, this module first concatenates $x^{link}_i$, $x^{heat}_i$, the MLP-processed $x^{freq}_i$, and $lm_{i,k}$. It then processes the concatenated features through MLP and GAT\cite{GAT_first} to generate the link representations $x^{l}_{i,k}$ for $l_i$ in the $k$-th iteration. $lm_{i, k}$ represents the conditional representation of $l_i$ derived from $R_{k-1}$ under the correctional boosting approach, which will be elaborated in the next section.

\textbf{Multi-Scenario Network}: This network calculates the link cost using $x^u$ and $x^l_{i,k}$ with DSFNet \cite{DSFnet2025}, a scenario-specific network designed for route recommendation, which enhances multi-scenario capabilities by decomposing parameters into several scenario-specific learners. The formula for calculating the link cost is as follows:

\begin{equation}
c_{i, k} = \text{ReLU}(\text{DFSNet}(x^u, x^{l}_{i,k})), \\
\label{eq: link cost model 3}
\end{equation} 
where ReLU is applied to constrain the cost to positive values, which is necessary for routing with Dijkstra algorithm. Additionally, discrete features are embedded and continuous features are standardized prior to being input into the LCM.

\subsection{Route Generation}
\textbf{Training}. Since the Dijkstra routing algorithm is non-differentiable, we use a pre-generated route set $R$ to make training feasible, reducing route generation to calculating the selection probabilities for each route. In the $k$-th iteration, after computing the cost $c_{i,k}$ for each link in graph $G'$, the cost of each route in $R$ is obtained as the sum of the costs of the links forming the route. Taking the negation of these costs, the softmax function is applied to derive the probability distribution $P_k$ over all routes in $R$:
\begin{equation}
P_k = \text{Softmax}\left(- \Big[ \sum_{l_i \in r_j} c_{i,k} \Big]_{r_j \in R} \right),
\label{eq: route p}
\end{equation}
where $[ \dots ]_{r_j \in R}$ denotes a vector containing the costs of all routes in $R$. The route with the highest probability in $P_k$ is selected as the route generated in the $k$-th iteration, denoted as $r_k$. Since routing algorithms prioritize lower costs, the negative sign is applied to ensure that maximizing $P$ is equivalent to minimizing costs.

\textbf{Inference}. In the $k$-th iteration of the inference phase, based on $O$, $D$, $G'$, and the computed link costs, Bidirectional Dijkstra Routing is invoked to find the optimal route $r_k$.

\subsection{Correctional Boosting}
We propose the correctional boosting method to iteratively update link costs and generate routes, achieving both personalization and diversity.
Revisiting the correctional boosting objective \cref{eq: obj_f}, $R_{k-1}$ serves as a conditional dependency for route probability computation in the $k$-th iteration. To incorporate the information of $R_{k-1}$ into the model, we design a link memory mechanism.

\subsubsection{Link Memory}
Link Memory $lm$ is a vector of length $K$, where the $k$-th dimension indicates whether a link belongs to the route $r_k$ generated in the $k$-th iteration. Let $lm_{i,k}$ represent the memory vector of link $i$ in the $k$-th iteration, and its update formula is given as follows:

\begin{equation}
lm_{i,k} = lm_{i,k-1} + \mathbf{e}_k \cdot \mathbf{1}[l_i \in r_k], \\
\label{eq: link mem}
\end{equation} 
where $\mathbf{e}_k$ is the $K$-dimensional one-hot vector where only the $k$-th position is 1, and $\mathbf{1}[i \in r_k]$ is the indicator function, which equals 1 if $l_i$ belongs to the route $r_k$; otherwise, it equals 0. $lm_{i,k}$ is used as a conditional representation in the LCM, as shown in \cref{fig: framework}(b).

From \cref{eq: obj_f}, we know that in the $k$-th iteration, the samples requiring further optimization have not yet generated the route $r_*$. In this context, $lm$ serves to inform the model which links have been selected and how many times they have been selected. This encourages the model to reduce the likelihood of frequently selected links appearing in subsequent iterations.

\subsubsection{Correctional Boosting Loss}
Based on equations \eqref{eq: probForm}-\eqref{eq: link mem}, the correctional boosting loss during the training phase is defined as:

\begin{gather}
\mathcal{L}_k = -\sum_{d\in\Omega}(Cov_{*, d}-Cov_{k-1,d})\cdot Y_d\log P_{k,d}, \notag \\
\mathcal{L} = \sum_{k=1}^K \mathcal{L}_k, 
\label{eq: all loss}
\end{gather}
where $d$ is a sample in the dataset $\Omega$, and $Y$ is the one-hot label corresponding to the route set $R$, with the value of $r_*$ set to 1 and all other values set to 0. The total loss is the sum of the per-iteration losses $\mathcal{L}_k$. The loss function can essentially be regarded as a weighted cross-entropy loss, where the weights are designed to encourage diversity. 

\subsection{Computational Efficiency}
To improve training and inference efficiency, we introduce the Skeleton-to-Capillary approach to downscale request-level road networks. During training, a route sampling method is applied to reduce the size of $R$. For online inference, we remove GAT from the LCM to improve computational efficiency.

\subsubsection{Skeleton-to-capillary}
\begin{figure}[t]
  \centering
  \includegraphics[width=\linewidth]{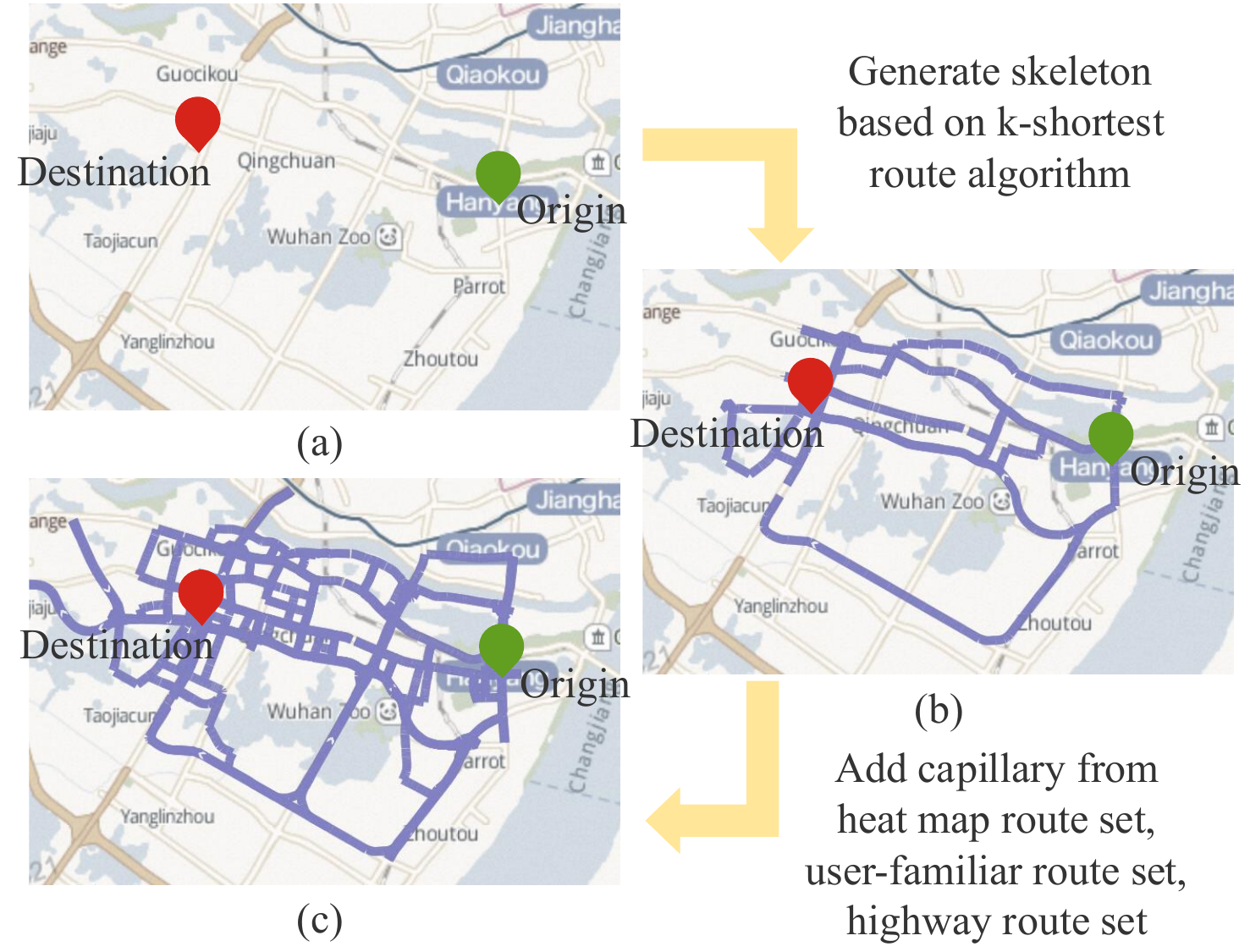}
  \caption{Skeleton-to-capillary (STC).}
  \label{fig: sub_network}
\end{figure}

The Skeleton-to-Capillary method generates a sub-network for each request. As illustrated in \cref{fig: sub_network}(a), given the complete road network along with $O$ and $D$, STC first applies the method from \cite{subnetwork2010} to construct an initial skeleton sub-network, as shown in \cref{fig: sub_network}(b). To further enhance the coverage of potential links that users may traverse, STC analyzes each interaction pair $(v_i, v_j)$ on the skeleton sub-network, searching for high-heat links, user familiar links, and highway links between $v_i$ and $v_j$ based on trajectory data from the past three months. These links are added to the sub-network, forming the capillary structure. STC significantly reduces the network scale (remaining $5\%$) while maintaining high-quality links. 

The process of STC method is as follows:

\textbf{Step 1} Generate the shortest route $r_*$ between the OD points in the whole network, and randomly generate a large number of feasible routes through traversal, which are stored in $\mathcal{R}$;

\textbf{Step 2} For any route $r_i \in \mathcal{R}$, calculate the local optimal proportion coefficient $l_i$, the similarity coefficient $s_i$ and the detour coefficient $d_i$ compared to route $r_*$, where $l_i$ refers to the ratio of the length of the maximum local optimal sub route of $r_i$ to the total length of $r_i$;
\begin{equation} \label{eq: thresholds1}
    l_i=\frac{\max\limits_{r\in sub(r_i),r\in optimal}{|r|}}{|r_i|}, 
\end{equation}
\begin{equation} \label{eq: thresholds2}
    s_i=\frac{|r_i\cap r_*|}{|r_*|}, 
\end{equation}
\begin{equation} \label{eq: thresholds3}
    d_i=\frac{|r_i|}{|r_*|}, 
\end{equation}
where $sub(r_i)$ is the set of all sub routes of $r_i$; $optimal$ is the set of all the optimal routes of any OD pair in the network.

\textbf{Step 3} Set three thresholds $l_0, s_0, d_0$, filter out the routes in $\mathcal{R}$ where $l_i<l_0$, or $s_i>s_0$, or $d_i>d_0$; the remaining routes form the skeleton network $\mathcal{N}_{ini}$;

\textbf{Step 4} Analyzes the user's travel scenarios (such as commuting, tourism, emergencies, etc.), and configures the corresponding expansion strategies according to the characteristics of the scenarios, such as the low cost strategy, high-speed priority strategy, etc., and then traverses all node pairs in $\mathcal{N}_{ini}$, for all routes connected to the node pair and not in network $\mathcal{N}_{ini}$, join $\mathcal{N}_{ini}$ according to different priorities according to the expansion strategies and get the sub-network $\mathcal{N}_{sub}$.

Each sample in the PRN dataset is constructed based on the STC method. 

\subsubsection{Route Sampling for Training}
The training phase requires a precomputed route set $R$, which ideally would include all possible routes in graph $G$. However, with up to 5000 links in $G$, the number of potential routes becomes astronomically large. To improve training efficiency, we generate 100 routes using a 6-dimensional Pareto link cost \cite{multipleobj2013shortest}, and select the following for training: the 10 highest-rated routes, 6 single-dimension optimal routes, and 10 random routes. The route ratings are determined using a pre-trained ranking model from a navigation application. 

Our sampling method (RS) is as follows:

\textbf{Step 1} Select the six most important dimensions(time, distance, cost, proportion of familiar roads, number of traffic lights, length of rough roads) and characterize each road as a vector $a$ containing these six dimensional features;

\textbf{Step 2} Each route is a six-dimensional vector $a$, that is, all routes represent points in a six-dimensional space, and then calculate the six-dimensional Pareto surface $\Gamma$ of all routes;

\textbf{Step 3} Select all routes on the Pareto surface $\Gamma$ into the sampling set $R$.

According to the sampling method, we get all the Pareto optimal solutions. For any remaining route $r_i\notin R$, there is at least one route $r_j\in R$, which satisfies $r_j$ exceeds $r_i$ in all six dimensions, to ensure the quality of the sampled route and the training effect. The route set $R$ in the PRN dataset is already pre-sampled using this approach.

\subsubsection{Incremental Online Inference}
To meet real-time requirements, we simplify LCM by removing the GAT module. Without GAT, link cost calculations become independent. In the $k$-th iteration, only links with updated $lm_{i,k}$ are recalculated. This reduces computational complexity from $N \times K$ to $N + |r| \times (K-1)$, where $|r|$, the average number of links per route, is much smaller than $N$. Experiments show that removing GAT significantly improves efficiency with minimal performance loss.

\subsection{Training Process}
The whole training process can be described as Algorithm 1.
\begin{algorithm}[ht!]
\caption{Training process}
\label{alg: framework}
\SetAlgoNlRelativeSize{-1}
\KwIn{Features $x^s, x^h, x^{link}, x^{heat},x^{freq}$}
\KwOut{Link cost model}
\textbf{Step 1: Framework Initial}

Generating the sub-network $\mathcal{N}_{sub}$ via STC method in section 4.3

Convert $\mathcal{N}_{sub}$ to dual graph $G'$ via method in section 3

Preprocessing features 

Initial $lm_1=(lm_{1,1},...,lm_{N,1})$ as the zero vector

Sampling routes from origin to destination

Calculate the coverage $Cov$ of each route via \cref{eq: probForm}

Calculate the one-hot Label $Y$

\textbf{Step 2: Route Planning}

\While {$Cov_{k-1}\neq Cov_*$}{
    Calculate the costs $c_{i,k}$ for each link via \cref{eq: link cost model 3}
    
    Find the route $r_{k}$ with the minimum total cost.

    Calculate the weight vector $Cov_*-Cov_{k}$ and update the link memory $lm_{k}$
}

\textbf{Step 3: Loss calculation}

Calculate the total loss $\mathcal{L}$ via \cref{eq: all loss}

\Return $\mathcal{L}$ to optimize
\end{algorithm}

\section{Experiment}
\subsection{Experiment Settings}
\subsubsection{Datasets and Model Parameter Settings} 
\textbf{Datasets}. We sampled $500K$ instances from the PRN dataset as the training set and an additional $50K$ instances as the test set. To better evaluate the performance of each method in different scenarios, we further process the test set into four categories: the full test set ($set_1$); samples with non-empty user familiarity frequency ($set_2$); samples with non-empty user historical sequences ($set_3$); and the union of $set_2$ and $set_3$ ($set_4$).

\textbf{Model Parameter}. The parameters of the MLP in the link representation module and multi-scenario network are $128\times 64\times32$,$128\times64\times32\times1$, respectively, with ELU as the activation function. GAT employs a single-layer attention mechanism.

\begin{table}[t]
  \caption{Compare the $Cov_1$(\%) and efficiency of different methods under the Single-Route Generation task.}
  \label{tab: single}
  \centering
  \begin{tabular}{lccccc}
    \toprule
    \textbf{Method} & $Set_1$ & $Set_2$ & $Set_3$ & $Set_4$ & RT(ms) \\
    \midrule
    ST       & 73.5 & 70.8 & 72.0 & 69.5 & 15.82 \\
    SD       & 70.2 & 66.6 & 68.4 & 65.2 & 17.73  \\
    MT       & 67.8 & 63.6 & 66.5 & 62.9 & 17.32 \\
    HF       & 33.8 & 77.1 & 40.0 & 76.5 & 17.85 \\
    GA       & 66.0 & 61.8 & 67.7 & 62.5 & 1765.8 \\
    \textbf{GenMRP} & \textbf{79.0} & \textbf{81.4} & \textbf{78.7} & \textbf{80.2} & 57.85  \\
    \bottomrule
  \end{tabular}
\end{table}

\subsubsection{Computational Resources}
\textbf{Training}. All models were trained using TensorFlow toolbox on a Tesla H20 GPU with 96GB memory. Our model typically converges in approximately 24 hours.

\textbf{Inference}. All models were inferred using INTEL(R) XEON(R) PLATINUM 8575CPU (3.2GHz).

\subsubsection{Baselines and Metrics} 
We evaluate GenMRP's performance on two tasks:

\textbf{Single-route generation}: This task evaluates the effectiveness of the optimal route generated by each method, using the coverage of the optimal route ($Cov_1$) as the evaluation metric. We compare GenMRP with five baseline methods: 
\underline{Shortest Time} (ST), \underline{Shortest Distance} (SD), 
\underline{Minimum Toll} (MT): Compute the route with minimal travel time, distance, and toll, respectively. \underline{Highest} \underline{Familiarity} (HF) \cite{finding2017prefer}: Enhances route generation by prioritizing links based on user historical trajectories. 
\underline{Generative Approach} (GA) \cite{RL2019personal, GA2023}: Uses a generative framework to update link costs and generate the next link iteratively.  

\textbf{Multi-route generation}: This task evaluates the effectiveness of generating multiple routes, with the maximum coverage ($Cov_K$) of the route set as the evaluation metric. We compare GenMRP with five baseline methods: 
\underline{K-shortest Time} (KST), 
\underline{K-shortest Distance} (KSD), 
\underline{K-minimum Toll} (KMT): Computes $K$ routes minimizing travel time, distance and toll, respectively. 
\underline{K-highest Familiarity} (KHF): Focuses on $K$ routes favoring familiar links based on user historical trajectories. 
\underline{2D Pareto} (2DP): Computes routes based on multi-objective optimization using travel time and distance. 

The first four baselines compute multiple routes via the alternative routes algorithm \cite{alternative2013route}, while the Pareto methods leverage multi-objective optimization frameworks \cite{multipleobj2013shortest}. In the experiments, $K$ is set to 3. Additionally, we use response time (RT) to evaluate the inference efficiency of each method.

\subsection{Experiment Results}
\subsubsection{Single-route Generation}
From \cref{tab: single}, GenMRP achieves the highest $Cov_1$ across the test datasets, outperforming baselines by over 5\% on test set 1, with even greater improvements on personalized test sets 2 to 4. This highlights that user route choices are personalized rather than purely objective-driven, and GenMRP effectively captures this preference.

\subsubsection{Multi-route generation}
\begin{table}[t]
  \caption{Comparison of $Cov_K$(\%) and efficiency across different methods in the Multi-Route Generation task.}
  \label{tab: multi}
  \begin{tabular}{lccccc}
    \toprule
    \textbf{Method} & $Set_1$ & $Set_2$ & $Set_3$ & $Set_4$ & RT(ms) \\
    \midrule
    KST & 79.4 & 76.8 & 78.5 & 76.1 & 18.5\\
    KSD & 76.5 & 73.1 & 75.2 & 72.0 & 22.6\\
    KMT & 72.1 & 67.9 & 71.1 & 67.7 & 22.1\\
    KHF & 35.0 & 80.0 & 41.7 & 79.7 & 21.0\\
     2DP & 75.6 & 74.6 & 74.4 & 73.8 & 49.7\\ 
    \textbf{GenMRP} & \textbf{84.0} & \textbf{85.9} & \textbf{84.6} & \textbf{85.6} & 173.5 \\ 
    \bottomrule
  \end{tabular}
\end{table}

The experimental results are shown in \cref{tab: multi}. GenMRP achieves significant improvements in $Cov_k$ across all test sets, with the largest gains observed on personalized test sets 2 to 4. It is worth noting that GenMRP has the highest RT; however, a more efficient model version will be selected based on additional experiments in Section 5.3.

\subsubsection{Diversity Analysis}

\begin{figure}[t]
  \centering
  \includegraphics[width=\linewidth]{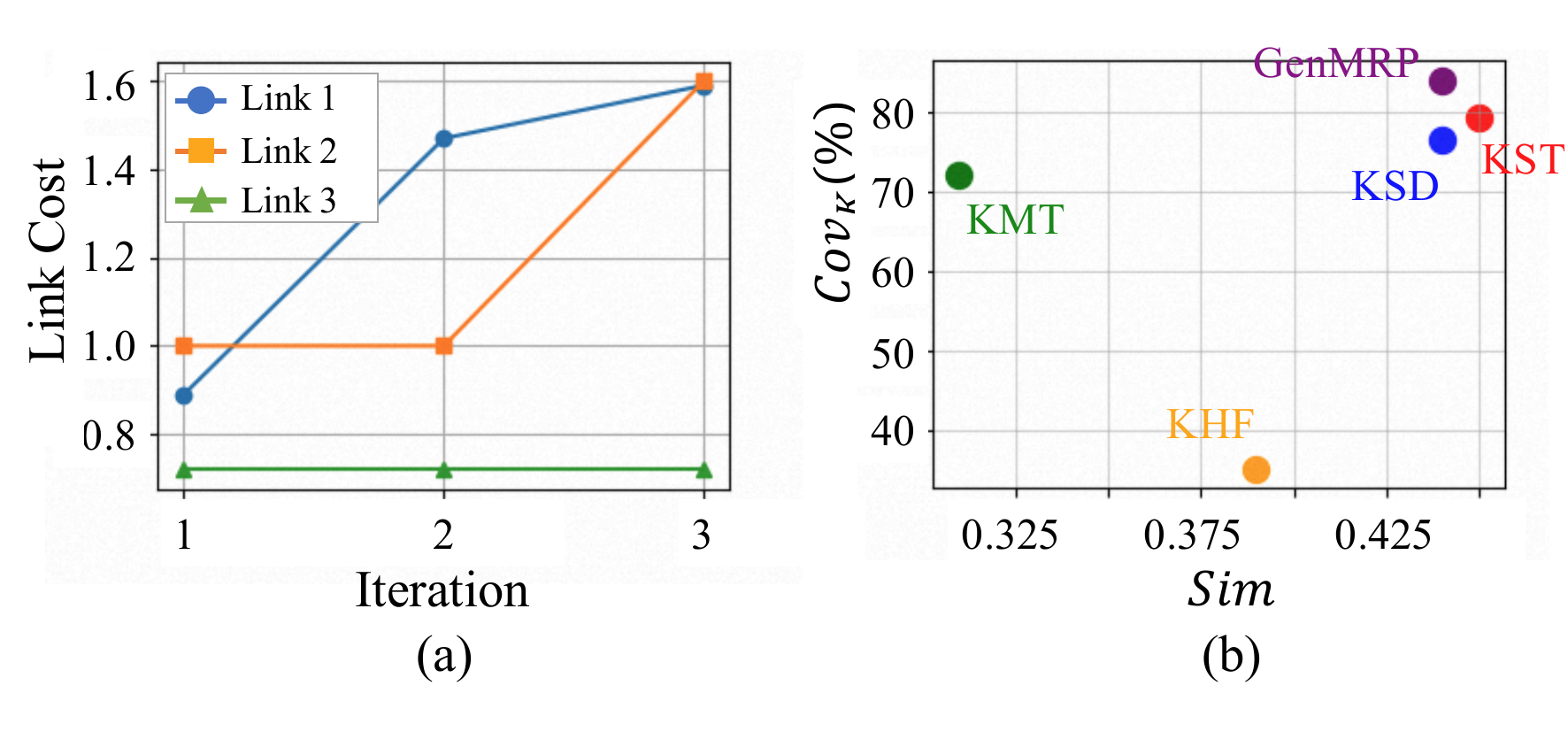}
  \caption{Diversity analysis.}
  \label{fig: cost change}
\end{figure}

\cref{fig: cost change} (b) depicts the relationship between route set similarity and $Cov_k$ across different methods. GenMRP achieves the highest $Cov_k$ with moderate similarity among routes. While the KMT method has the lowest similarity, its poor $Cov_k$ demonstrates that low similarity alone is insufficient to achieve meaningful diversity. The goal is to ensure diversity by generating routes that are both dissimilar and “useful” to users. The calculation method for route set similarity is as follows:

\begin{equation}
Sim = \frac{1}{\sum_{r_i,r_j\in R}1}\sum_{r_i,r_j\in R} \frac{|r_i\cap r_j|}{|r_i\cup r_j|}, \\
\label{eq: sim}
\end{equation} 

\cref{fig: cost change}(a) illustrates the cost changes for links that first appear in each of the three iterations (Link $i$ represents the link being generated for the first time in the $i$-th iteration). It can be observed that links appearing in earlier iterations experience increased costs in subsequent iterations, which reduces their probability of being selected again. This explains how the correctional boosting mechanism enables the model to balance personalization while maintaining diversity.

\subsection{Ablation Study}
 We performs ablation studies to evaluate the contribution of each component to GenMRP performance. Where (w/o w) represents the model that removes the weight $Cov_{*,d}-Cov_{k-1,d}$ in the loss function; (w/o u) represents the model that removes the user preference capture module; and (w/o GAT) represents the model without the GAT module. 
\begin{table}[t]
  \caption{Ablation study of model performance on $Cov_K(\%)$.}
  \label{tab: ablation}
  \begin{tabular}{lccccc}
    \toprule
    \textbf{Method} & $Set_1$ & $Set_2$ & $Set_3$ & $Set_4$ & RT(ms)\\
    \midrule
    w/o w & 83.9 & 85.3 & 84.3 & 84.7 & 173.5\\
    w/o u & 82.8 & 85.1 & 82.3 & 84.3 & 173.5 \\
    w/o GAT & 83.7 & 85.0 & 83.7 & 84.7 & \textbf{37.8} \\
    \textbf{GenMRP} & \textbf{84.0} & \textbf{85.9} & \textbf{84.6} & \textbf{85.6} & 173.5 \\ 
    \bottomrule
  \end{tabular}
\end{table}
As shown in \cref{tab: ablation}, ablating any of the three components leads to performance degradation, with the user preference capture module having the largest impact. More importantly, we observe that removing GAT and adopting incremental inference significantly reduces RT while maintaining acceptable performance loss. As a result, the GAT-free model is deployed online to meet real-time efficiency needs.

\subsection{Online Experiments}
We conducted online experiments on a real-world navigation app. The base online method combines KST, KSD, KMT, and KHF, augmented by numerous manual rules. In the A/B tests, we replaced KHF with GenMRP. The experiment ran for one week, with each group covering 30 million users. Alongside $Cov_K$, online evaluation metrics include Deviation Rate (DR), $Cov_{net}$, and $N_p$. DR quantifies the proportion of users who deviated from the recommended routes. $N_p$ measures the average number of Pareto-optimal routes in the overall route set, serving as an indicator of diversity. $Cov_{R}$ assesses the coverage of all links across generated routes against user trajectories. $Cov_{net}$ is calculated as $\tfrac{|r_u \cap (\bigcup_{R} L)|}{|\bigcup_{R} L|} $. As presented in \cref{tab: online result}, GenMRP achieves a $Cov_K$ improvement of over 0.5\%, representing the largest gain in the past year. Moreover, it shows advancements across other metrics. GenMRP has now been fully deployed online.

\begin{table}[t]
  \caption{Online experiment result.}
  \label{tab: online result}
  \begin{tabular}{lcccc}
    \toprule
    Method & $Cov_K$ & DR & $N_P$ & $Cov_{net}$\\
    \midrule
    Base & 91.10\%	& 15.33\% &  2.91 & 93.25\% \\
    \textbf{GenMRP} & \textbf{91.64\%} & \textbf{15.15\%}  & \textbf{3.10} & \textbf{93.92\%} \\ 
    \bottomrule
  \end{tabular}
\end{table}

\subsection{Case Study}
\begin{figure}[t]
  \centering
  \includegraphics[width=\linewidth]{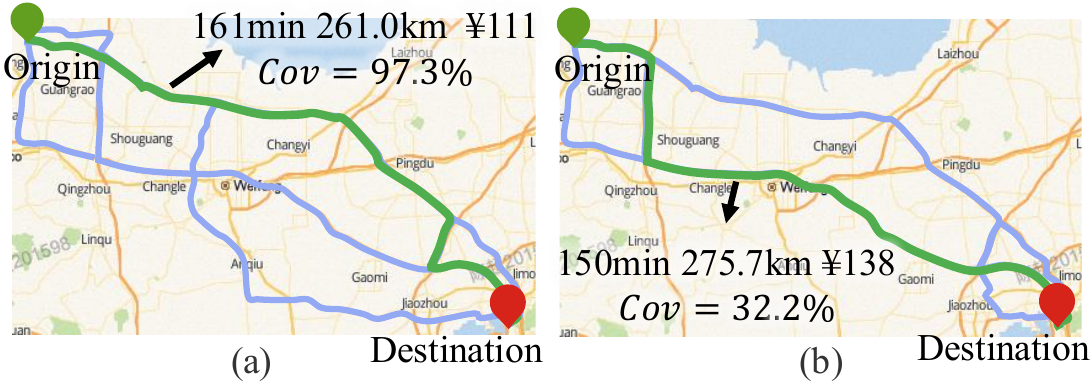}
  \caption{A real-world case.}
  \label{fig: case online}
\end{figure}

\cref{fig: case online} illustrates a real-world case, where (a) and (b) respectively show the route sets generated before and after the online deployment of GenMRP. The first route (in green) generated by GenMRP achieves higher coverage, while the diversity of the set of routes is significantly improved.

\section{Conclusion}
We present GenMRP, a generative framework for multi-route planning that addresses the challenges of personalization, diversity, and efficiency. GenMRP integrates contextual features,  user historical sequence, and various link features into a deep learning-powered Link Cost Model using DIN, GAT, and a Multi-Scenario Network. To enhance diversity, an iterative correctional boosting mechanism adaptively refines routes based on previously generated ones. To meet real-time industrial requirements, we introduce the Skeleton-to-Capillary (STC) approach and incremental inference, enabling dynamic sub-network construction and efficient updates. Moreover, we release PRN, the first request-level road network dataset for research. GenMRP has been fully deployed in a real-world navigation application.

\bibliographystyle{ACM-Reference-Format}
\bibliography{sample-base}

\clearpage
\appendix
\section{Feature description}
\label{Appendix:feature}
Context-aware feature $x^s$, which is a vector with a shape of $[10]$, including od coordinates, whether at night and other information, this feature generally describes the currently requested scene information; user history sequence $x^h$ is a vector in the shape of $[100, 32]$, which contains information such as the user's selection strategy (high-speed priority, time priority, cost priority, etc.) for the last 100 times and the characteristics of the user's route selection (route length, time, cost, etc.), which records the user's historical selection behavior and can learn the user's historical preference information; link static feature $x^{link}$ of the road network is a matrix with the shape of $[N, 50]$, $N$ is the number of links in the dual network, $50$ represents the number of feature dimensions of each link, including static information such as length, road grade, etc; dynamic characteristics of dual network (frequency features $x^{freq}$ and OD heat data $x^{heat}$), $x^{freq}$ is a tensor with the shape of $[N, 20, 7]$, $20$ represents that each link records the user's last $20$ requests through the link within three months, and $7$ is the number of feature dimensions of each historical request, including the time from the historical request to the current time, the distance between the historical OD points and the current OD points, etc., $x^{heat}$ is a tensor with the shape of $[N,2]$, which records the number of times each link in the road network in all users' requests with the same OD points within three months; route set $R$ is a matrix with the shape of $[M,|r|]$, $M$ represents the number of routes, $|r|$ is the number of links for each route, and $Cov$ is a vector with the shape of $[M,1]$, which records the coverage of each route compared with the user's actual route. 

\end{document}